%% file: root.tex
\documentclass[letterpaper, 10 pt, conference]{ieeeconf}  

\IEEEoverridecommandlockouts                              

\usepackage{times}
\usepackage{epsfig}
\usepackage{graphicx}
\usepackage{amsmath}
\usepackage{amssymb}
\usepackage{microtype}

\usepackage{float}
\usepackage{bm}
\usepackage{epstopdf}
\usepackage{comment}

\usepackage{multirow}
\usepackage[table,xcdraw]{xcolor}
\usepackage{multirow}
\usepackage[table,xcdraw]{xcolor}
\usepackage[normalem]{ulem}
\useunder{\uline}{\ul}{}
\usepackage{subfig}
\usepackage{comment}
\usepackage{cite} 
\usepackage{hyperref}
\usepackage[stable]{footmisc}

\newcommand{\etal}{et al. }
\newcommand{\norm}[1]{\left\lVert #1 \right\rVert}

\title{\LARGE \bf
Towards Anomaly Detection in Dashcam Videos  
}

\author{Sanjay Haresh$^*$~~~~~Sateesh Kumar$^*$~~~~~M. Zeeshan Zia~~~~~Quoc-Huy Tran
\thanks{$^*$ indicates equal contribution.}
\thanks{All authors are with Retrocausal, Inc., Seattle, WA 98105, USA.
        {\tt\small \{sanjay,sateesh,zeeshan,huy\}@retrocausal.ai}}
}

\begin{document}

\maketitle
\thispagestyle{empty}
\pagestyle{empty}

\input{abstract}

\input{introduction}

\input{relatedwork}

\input{approach}

\input{dataset}

\input{experiments}

\input{conclusion}

{\small
\bibliographystyle{IEEEtran}
\bibliography{reference}
}

\end{document}

%% file: abstract.tex
\begin{abstract}
Inexpensive sensing and computation, as well as insurance innovations, have made smart dashboard cameras ubiquitous. Increasingly, simple model-driven computer vision algorithms focused on lane departures or safe following distances are finding their way into these devices. Unfortunately, the long-tailed distribution of road hazards means that these hand-crafted pipelines are inadequate for driver safety systems. We propose to apply data-driven anomaly detection ideas from deep learning to dashcam videos, which hold the promise of bridging this gap. Unfortunately, there exists almost no literature applying anomaly understanding to moving cameras, and correspondingly there is also a lack of relevant datasets. To counter this issue, we present a large and diverse dataset of truck dashcam videos, namely \emph{RetroTrucks}, that includes normal and anomalous driving scenes. We apply: (i) one-class classification loss and (ii) reconstruction-based loss, for anomaly detection on RetroTrucks as well as on existing static-camera datasets. We introduce formulations for modeling object interactions in this context as priors. Our experiments indicate that our dataset is indeed more challenging than standard anomaly detection datasets, and previous anomaly detection methods do not perform well here out-of-the-box. In addition, we share insights into the behavior of these two important families of anomaly detection approaches on dashcam data.

\end{abstract}

%% file: introduction.tex
\section{Introduction} 
\label{sec:introduction} 


Smart dashboard cameras have become ubiquitous in recent years, and several model-driven accident warning systems have been proposed for these cameras, e.g. lane departure warning (LDW) and forward collision warning (FCW) \cite{mukhtar2015vehicle, narote2018review}. Unfortunately, these approaches are limited to specific modes of accidents, whereas the problem is long-tailed, i.e. most anomalies cannot be explicitly modeled. Yet other approaches, especially in the advanced driver-assistant system (ADAS) literature, require sensors that are significantly more expensive than a monocular camera \cite{liu2017radar, schneider2019lidar}. In this paper, we study the problem of detecting anomalies on road scenes from dashcam videos in a purely data-driven setting. Our objective is to explore machine learning models that can prevent a broader class of accidents than those addressed by explicit models. 

We have seen great progress in video understanding tasks such as action recognition and activity detection in recent years. Much of this progress can be attributed to large-scale datasets \cite{karpathy2014large, carreira2017quo, idrees2017thumos}. Unfortunately, anomaly detection in videos remains a sparsely explored problem. To our knowledge, existing video anomaly detection literature is limited to surveillance in static-camera scenes \cite{li2013anomaly, lu2013abnormal, luo2017revisit, Sultani2018RealWorldAD}. Often the anomalies of interest are characterized by relatively simple visual phenomena, such as sudden motion in a region of the video frame, or by visual artefacts or objects not present in the training data. In contrast, driving scenes exhibit continuous motion, and anomalies are often characterized by complex interactions between traffic participants. Thus, we propose a large and diverse dataset of truck dashcam videos curated from YouTube, that we name \emph{RetroTrucks}. Our dataset includes a variety of driving scenes including normal driving, collisions, and near-misses. Moreover, truck videos lend a novel viewpoint which has not been explored in the literature. We choose truck examples, because dashcams have an even greater penetration in commercial vehicles than private ones. We expect this dataset to present new challenges to the anomaly detection community, and draw their attention towards traffic accident understanding.


We note that it is significantly easier to collect normal driving videos than accident ones. We leverage this imbalance by emphasizing approaches that do not require anomalous exemplars for training. In particular, we explore two families of approaches: (i) one-class classification and (ii) reconstruction-based. One-class classification \cite{khan2014one, ruff2018deep, perera2019learning} refers to methods which learn a manifold for normal data while constraining the manifold to be as compact as possible. At test time, any data mapped outside the learned manifold are classified as anomalous. In particular, we use 3D convolutional neural networks (CNNs) to learn a manifold for normal video clips. Reconstruction-based approaches \cite{an2015variational, zhao2017spatio, liu2018future, gong2019memorizing} use autoencoders to learn to reconstruct the input data through a bottleneck representation. They are trained exclusively on normal data and poor reconstruction of the input data at test time is used as a cue to detect anomalies. More specifically, we use 3D convolutional autoencoders for learning reconstruction of normal video clips. Furthermore, we inject object interaction priors into the above approaches to model accidents which are often caused by collisions between traffic participants. We evaluate this novel idea in both one-class classification and reconstruction-based settings.

We find that one-class classification methods fail to perform anomaly detection reliably in driving scenarios. 
In addition, reconstruction-based approaches outperform one-class classification ones, however, they fail to replicate their performance seen on standard datasets \cite{li2013anomaly, luo2017revisit} to our dataset. We study failure cases and show that reconstruction-based approaches, although great at detecting anomalies characterized by novel visual artefacts, fail to detect anomalies involving complex interactions of objects. We discuss these observations in details in Sec. \ref{sec:experiments}. In summary, our contributions include:
\begin{itemize}
\item We propose different data-driven approaches for anomaly detection in dashcam videos, including modeling object interactions and exploring motion features.
\item We contribute RetroTrucks --- a new dataset for dashcam anomaly detection, which is useful for tasks such as traffic accident detection, ADAS, and road scene understanding in general~\cite{dhiman2016continuous,li2017deep}.
\item Our evaluation reveals insights on the performance of various anomaly detection methods on static-camera datasets \cite{li2013anomaly, luo2017revisit} and the proposed dashcam dataset.
 
\end{itemize}


%% file: relatedwork.tex
\section{Related Work}
\label{sec:relatedwork}

\noindent \textbf{Model-Driven Anomaly Detection:} Model-driven methods for anomaly detection and ADAS have garnered great research interest. Song \etal \cite{song2017fcwstereo} proposed a stereo vision based system for lane detection and forward collision warning, whereas Liu \etal \cite{liu2017radar} used radars to detect vehicles in the blind spot of the ego-car and generate warnings to avoid collisions. Recently, Matousek \etal \cite{matousek2019detecting} proposed to model driving behaviors using neural networks to detect accidents. Similarly, Fang \etal \cite{fang2019dada} proposed to use gaze estimation as a proxy for driver attention to detect accidents. Note that all of the above methods work for specific cases of anomalies but may fail to detect others.

\noindent \textbf{One-Class Classification Approaches:} These methods train machine learning models on an one-class classification objective, i.e. the models learn a manifold for the normal class only, as opposed to methods learning a hyperplane to distinguish between the two classes, i.e. normal and anomalous. These methods have a long history in classical machine learning \cite{khan2014one} but have only recently been adopted to deep learning. An end-to-end support vector data description (SVDD) objective for deep neural networks was first introduced in \cite{ruff2018deep}, where the network was trained to map normal data into a hypersphere and simultaneously minimize the volume of the hypersphere. The distance from the center of the hypersphere represents the anomaly score. Perera \etal \cite{perera2019learning}, on the other hand, used deep neural networks as feature extractors only and fed the extracted features to a classical one-class classifier. 

\noindent \textbf{Reconstruction-Based Approaches:} These methods learn to reconstruct from a compact representation of normal data and use poor reconstruction of the input data at test time as a cue to detect anomalous examples. The underlying assumption is that the representation capability of the learned models is so adjusted that they can only explain the variation in normal data and therefore fail to accurately reconstruct the anomalous examples. An and Cho \cite{an2015variational} were the first to use autoencoders for reconstruction-based anomaly detection in images. Zhao \etal \cite{zhao2017spatio} used 3D CNNs to simultaneously reconstruct the input frames and predict the future frames for anomaly detection in videos. Liu \etal \cite{liu2018future} on the other hand only predicted the future frames from the input frames for detecting anomalous events. Gong \etal \cite{gong2019memorizing}, however, noted that autoencoders generalize well even to unseen objects at test time and therefore introduced a memory module to regularize the representation capacity of the learned autoencoders. Reconstruction-based approaches, although effective, can model only ``visual'' aspects of the scenes and cannot exploit the more subtle contextual cues for real-world anomaly detection. We show how reconstruction-based approaches perform worse in such scenarios in Sec. \ref{sec:experiments}.  


\noindent \textbf{Weakly-Supervised Approaches:} Sultani \etal \cite{Sultani2018RealWorldAD} introduced a large-scale dataset for real-world anomaly detection, with temporal annotations for anomalous videos. Since their method used both normal and anomalous examples at training, it was a departure from the above approaches which require only normal examples for training, i.e. one-class classification and reconstruction-based. We emphasize that normal driving data on roads are several orders of magnitude more frequent than accident data, which is why we focus on approaches that require only normal exemplars for training. Yet, we provide additional anomalous videos with temporal annotations in our dataset to facilitate research in weakly-supervised approaches.

%% file: approach.tex

\section{Data-Driven Anomaly Detection for Dashcam Videos}
\label{sec:approach}

\begin{figure}[!t]
\centering
\subfloat[][]{\includegraphics[scale=0.33, trim = 0mm 0mm 110mm 40mm, clip, angle=270]{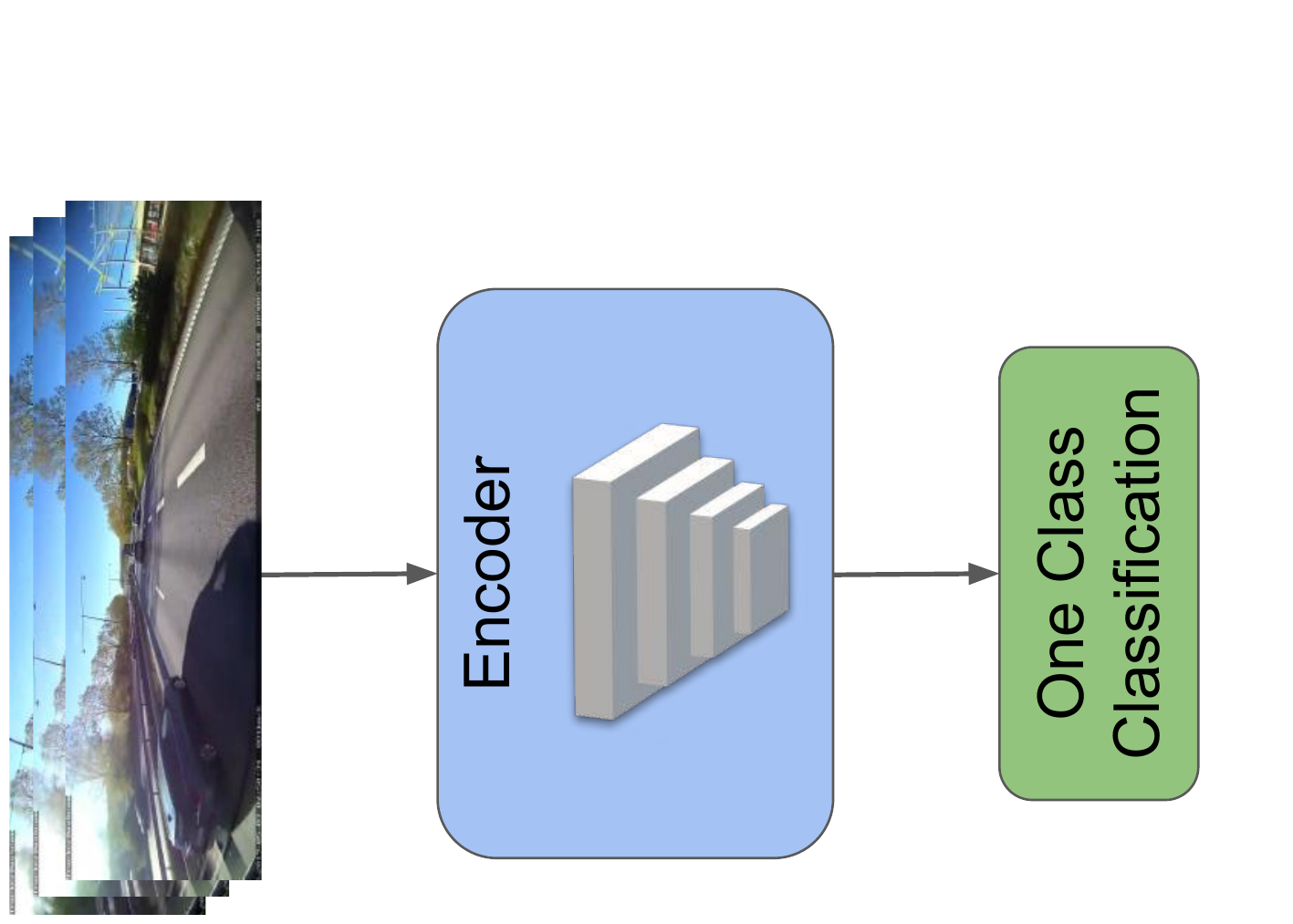}} 
\subfloat[][]{\includegraphics[scale=0.33, trim = 0mm 0mm 110mm 0mm, clip, angle=270]{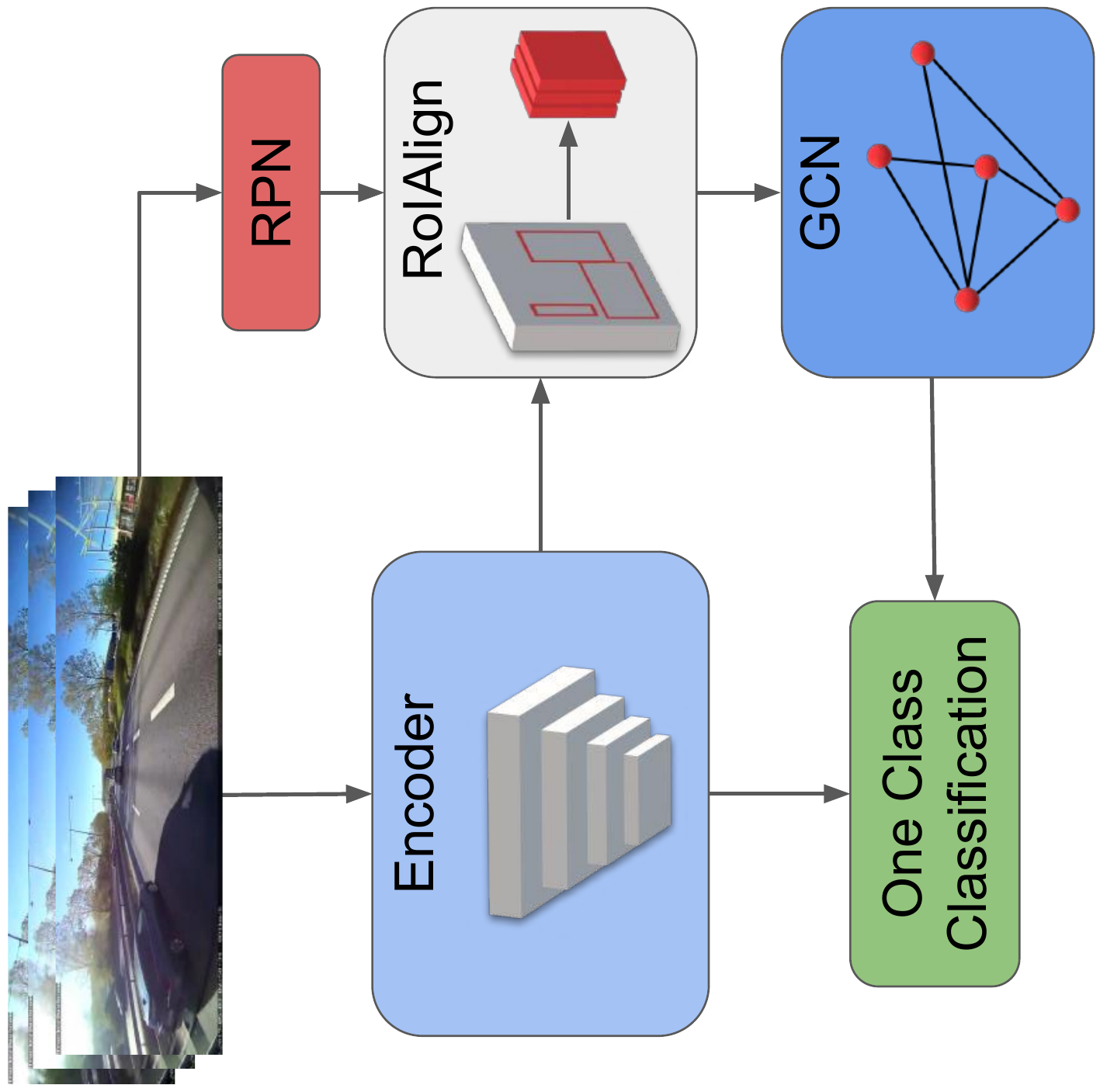}}
\vspace{-0.2cm}
\caption{Overview of one-class classification approaches. In (a), we train a 3D CNN with one-class classification objective on normal video clips. In (b), we augment (a) with object interaction reasoning by adding RPN and RoIAlign modules for detecting objects and extracting object features respectively, and a GCN module for reasoning about temporal object dependencies.}
\vspace{-0.5cm}
\label{fig:architecture_oc}
\end{figure}

We first explore two approaches for anomaly detection in dashcam videos: (i) one-class classification and (ii) reconstruction-based in Secs. \ref{sec:OC-SVDD} and \ref{sec:rec_ae} respectively. Next, we propose to incorporate object interaction priors, using graph convolutional networks (GCNs), in both of the above approaches in Sec. \ref{sec:gcn}. Figs. \ref{fig:architecture_oc} and \ref{fig:architecture_rec} provide an overview of our one-class classification and reconstruction-based approaches respectively.


\subsection{One-Class Classification Approach}
\label{sec:OC-SVDD}

One-class classification approach learns a manifold from normal data and any sample mapped outside the manifold at test time is classified as anomalous. In particular, we train a 3D CNN to encode normal video clips to a hypersphere while minimizing its volume. Intuitively, the latter acts as a regularization which forces the network to learn the minimal variation in normal clips. Thus, it will not be able to explain the large variation in anomalous clips, which will then be encoded outside the hypersphere. We employ the one-class deep SVDD objective from \cite{ruff2018deep}.

Formally, let $F(x; W) : \mathbb{R}^{X} \xrightarrow{} \mathbb{R}^{Z} $, implemented by a 3D CNN with weights $W$, map the input $x$ from the video clip space $\mathbb{R}^{X}$ to a point in the feature space $\mathbb{R}^{Z}$. Here, $X = T \times H\times W\times C$ is the dimension of the $T$-frame input clip $x$, constructed by sampling $T$ consecutive frames from a video, with each frame having a height $H$, width $W$, and $C$ channels. In addition, the feature dimension $Z$ is much smaller than the input dimension $X$. To minimize the volume of the hypersphere enclosing the normal clips encoded in the feature space, the one-class deep SVDD objective is defined as below:

\vspace{-0.2cm}
\begin{equation}
    \min_{\mathbb{W}} \frac{1}{N} \sum_{i=1}^{N} \norm{F(x_{i}; W) - c}^{2} + \frac{\lambda}{2} \norm{W}_{F}^{2}, 
\end{equation}
where $c \in \mathbb{R}^{Z}$ is the center of the hypersphere. The first term penalizes the $L_2$ distance between the feature point encoding the input clip and the center of the hypersphere, which encourages the network to encode the normal clips to a hypersphere that is as compact as possible. The second term is a standard regularization on the weights of the network, controlled by the parameter $\lambda$. At test time, for any $x \in \mathbb{R}^{X}$, the anomaly score can be computed as the distance between the feature point $F(x; W)$ and the center $c$ as $\norm{F(x; W) - c}^{2}$.

    

\subsection{Reconstruction-Based Approach}
\label{sec:rec_ae}

Here, we use reconstruction as a proxy task to perform anomaly detection. Intuitively, the idea is that an autoencoder trained on normal clips will be able to reconstruct normal scenes accurately but will fail on anomalous scenes due to the change in the data distribution. The autoencoder consists of two networks, i.e. an encoder and a decoder. The encoder takes as input a video clip and generates a bottleneck representation. The decoder then takes this bottleneck representation as input and reconstructs the video clip. We describe each component below:

\begin{figure}[!t]
\centering
\subfloat[][]{\includegraphics[scale=0.33, trim = 0mm 0mm 50mm 40mm, clip, angle=270]{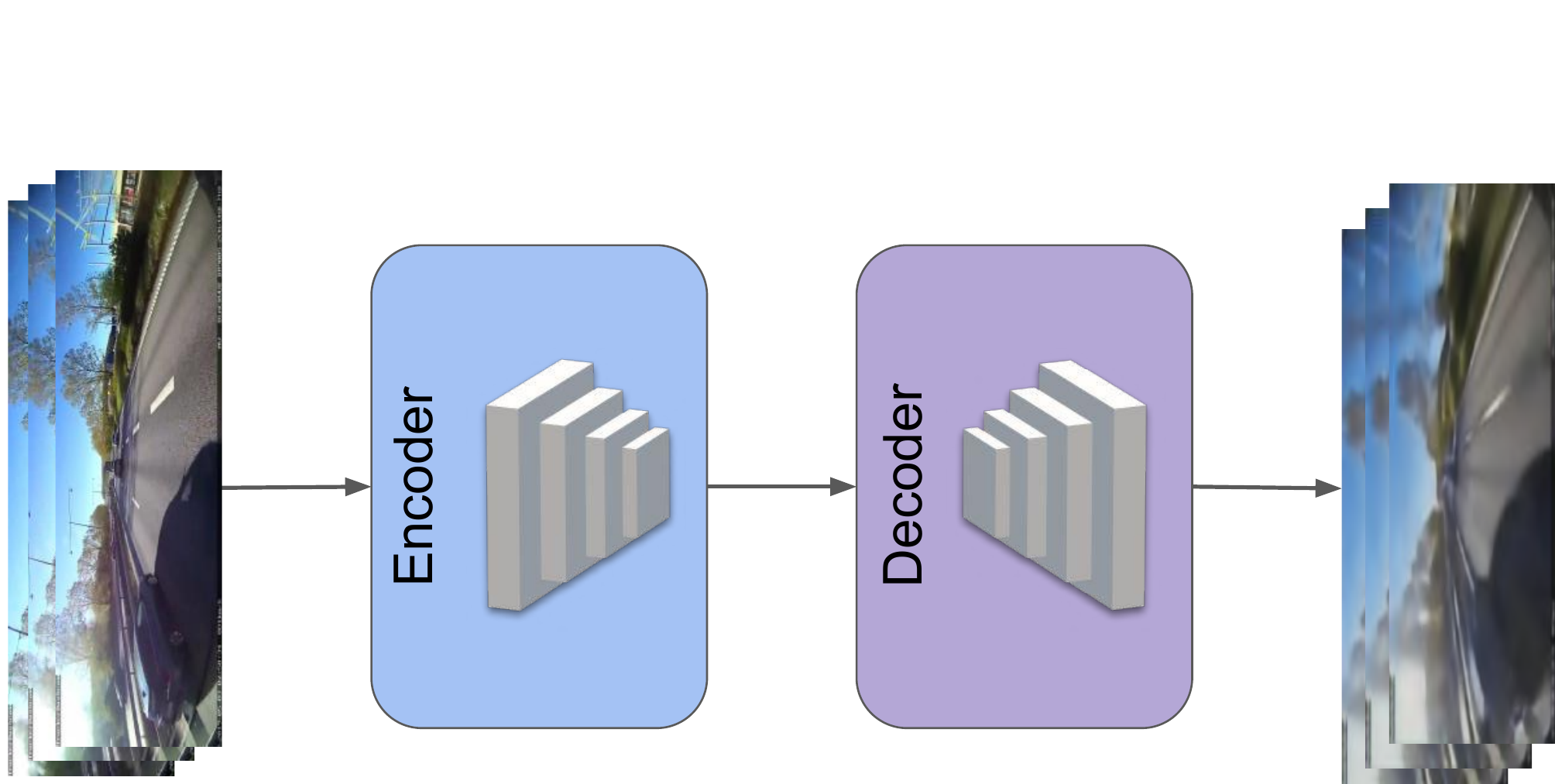}} 
\subfloat[][]{\includegraphics[scale=0.33, trim = 0mm 0mm 50mm 0mm, clip, angle=270]{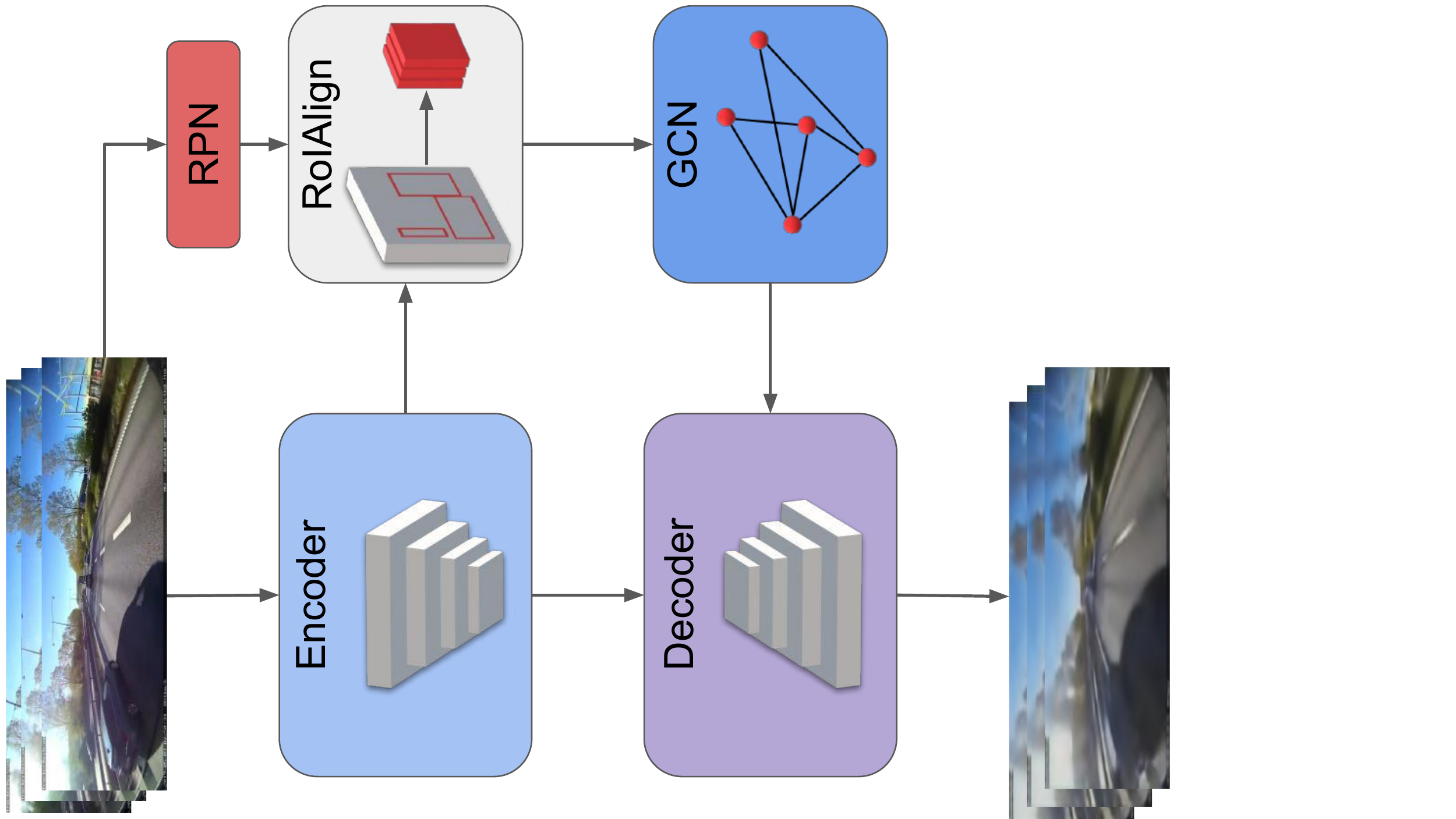}}
\vspace{-0.2cm}
\caption{Overview of reconstruction-based approaches. In (a), we use a 3D convolutional autoencoder to learn reconstruction of normal clips and use the reconstruction error as a score for anomaly detection at test time. In (b), we augment (a) with object interaction reasoning by adding RPN and RoIAlign modules for detecting objects and extracting object features respectively, and a GCN module for reasoning about temporal object dependencies.}
\vspace{-0.5cm}
\label{fig:architecture_rec}
\end{figure}

\noindent \textbf{Encoder:} Let $F_{e}(x; W_{e}) : \mathbb{R}^{X} \xrightarrow{} \mathbb{R}^{Z}$ be the encoder which encodes an input clip $x \in \mathbb{R}^{X}$ to a bottleneck representation in the feature space $\mathbb{R}^{Z}$. Next, instead of applying the one-class classification objective as in Sec. \ref{sec:OC-SVDD}, we reconstruct the input clip using the below decoder. 

\noindent \textbf{Decoder:} The decoder $F_{d}(x; W_{d}) : \mathbb{R}^{Z} \xrightarrow{} \mathbb{R}^{X}$ takes the bottleneck representation in the feature space $\mathbb{R}^{Z}$ and reconstructs the input clip ${x} \in \mathbb{R}^{X}$. 

In summary, we have:
\vspace{-0.4cm}
\begin{equation}
\begin{split}
    h = F_{e}(x; W_{e}),\hspace{0.2cm}
    \hat{x} = F_{d}(h; W_{d}),
\end{split}
\end{equation}
where $h$ is the bottleneck representation and $\hat{x}$ is the reconstruction of ${x}$ by the network. We minimize the $L_2$ loss between the input $x$ and the reconstructed $\hat{x}$ to train the autoencoder as follows:

\vspace{-0.2cm}
\begin{equation}
\label{eq:euclidean}
    L = \frac{1}{N} \sum_{i=1}^{N} \norm{x_{i} - \hat{x}_i}^{2} 
\end{equation}

\subsection{Modeling Object Interactions}
\label{sec:gcn}

Most real-world anomalies are due to unusual interactions among objects. Therefore, we seek to infuse object interaction priors directly into the above networks in Secs. \ref{sec:OC-SVDD} and \ref{sec:rec_ae}. Inspired by the recent success of \cite{wang2018videos} in action recognition, we first detect objects and extract object features in individual frames using region proposal network (RPN) and RoIAlign modules \cite{ren2015faster,he2017mask} respectively, and then build a similarity graph which is fed to a graph convolutional network (GCN) module so that it can reason about object dependencies across time. The idea is that by giving the network strong priors in the form of a graph encoding object interactions over time, the network should be able to learn the normal object interaction modes. It would then be able to distinguish between object interactions in normal and anomalous scenes.

We use the same encoder and/or decoder as in Secs. \ref{sec:OC-SVDD} and \ref{sec:rec_ae}. Let us denote the dimension of the bottleneck representation output by the encoder as $Z = T' \times H' \times W' \times d$, where $T'$ is the number of feature frames, and $H'$, $W'$, and $d$ are the height, width, and number of channels respectively for each feature frame.

\noindent \textbf{Region Proposal Network (RPN) and RoIAlign:} Along with the encoder, an RPN module \cite{ren2015faster} is used to extract $M$ object proposals for each feature frame in the bottleneck representation. Given the object proposals, we use RoIAlign \cite{he2017mask} to extract object features with dimensions $3 \times 3 \times d$ for each object proposal. We then use max pooling to get a $1 \times 1 \times d$ feature vector for each object proposal. Since there are $T'$ feature frames and $M$ object proposals for each feature frame, we get the total object features with dimensions $(T' \times M) \times d$. 

\noindent \textbf{Graph Convolutional Network (GCN):} Given the object proposals with corresponding object features, we construct a similarity graph where each object proposal is treated as a node. The graph will have strong edges for object proposals that are visually similar or highly correlated for normal scenes. We then use a GCN module \cite{kipf2016semi} to reason over the similarity graph. A GCN is a natural fit for this task as it can account for an arbitrarily defined neighbourhood as opposed to a CNN which works over a fixed locality. Formally, let $P = \{ p_1, p_2, \dots, p_K \}$ (with $K = T' \times M$) be the set of object features for the object proposals generated by the above RPN and RoIAlign modules. The similarity among the object proposals can then be defined as below:

\vspace{-0.1cm}
\begin{equation}
    S(p_i, p_j) = \phi(p_i)\phi'(p_j),
\end{equation}
where $\phi(.)$ and $\phi'(.)$ are single-layer networks. Next, we perform softmax normalization to get the normalized similarity as follows:

\begin{equation}
\label{eq:sim_graph}
    G_{i,j}^{sim} = \frac{ \exp S(p_i, p_j) }{\sum_{j} S(p_i, p_j)}.
\end{equation}

The graph is then fed into a two-layer GCN. This allows the network to reason about object dependencies across time. The output from GCN is then combined with the output of the encoder. Overall, we have:

\vspace{-0.2cm}
\begin{equation}
\begin{gathered}
    h_{enc} = F_{e}(x; W_{e}), \quad h_{rpn} = F_{rpn}(x; W_{rpn}), \\
    h_{gcn} = GCN(h_{rpn}),~~~ 
    h_{rich} = [h_{enc}, h_{gcn}].
\end{gathered}
\end{equation}

The combined representation, $h_{rich}$, is then used in the one-class classification loss or provided to the decoder for reconstructing the input clip as in Figs. \ref{fig:architecture_oc}(b) and \ref{fig:architecture_rec}(b) respectively.

%% file: dataset.tex
\section{\emph{RetroTrucks} --- A New Dataset for Dashcam Anomaly Detection}
\label{sec:dataset}

\subsection{Previous Datasets}
Most existing datasets are aimed at understanding anomalies in surveillance videos. Li \etal \cite{li2013anomaly} proposed two datasets of video recordings of pedestrians walking on UCSD campus, i.e. \emph{UCSD Ped1} with 70 videos and \emph{UCSD Ped2} with 28 videos. The anomalies in these videos are characterized by the presence of non-pedestrian entities such as cars, trucks, etc. \emph{Avenue} dataset \cite{lu2013abnormal} consists of 37 two-minute videos captured from a static camera in a fixed scene. The anomalies include running, throwing waste, etc. However, all the videos are captured from one fixed camera position. To introduce more variation in the data, Lu \etal \cite{luo2017revisit} proposed \emph{ShanghaiTech}, which includes 13 different scenes. The anomalies, however, still mostly include appearance of non-pedestrian entities like cyclists, skaters, etc. whereas real-world anomalies are characterized by context rather than visual aspects of entities. Sultani \etal \cite{Sultani2018RealWorldAD} proposed a large-scale dataset, i.e. \emph{UCF-Crime}, to alleviate this issue. UCF-Crime includes 1900 videos collected from the Internet. However, their dataset only includes static-camera scenes. Recently, Herzig \etal \cite{herzig2019spatio} introduced a dashcam dataset for accident recognition, which predicts a single label (i.e. normal or anomalous) for the entire input video, as opposed to our task of accident detection, which temporally localizes the accident frames in the input video. Chan \etal \cite{Chan16accv} proposed a dataset of dashcam videos for accident prediction. However, their dataset consists of only 620 video clips (each only 5 seconds long) and the accidents do not involve the ego-car. More recently, Che \etal \cite{che2019d} introduced a new dashcam dataset, i.e. D$^2$-City, however, it focuses on general road scene understanding and hence contains very few accident videos for anomaly detection.


\begin{figure}
\scriptsize
\centering
\setlength{\tabcolsep}{1.5pt}
\begin{tabular}{cc}
\subfloat[][]{\includegraphics[width=0.48\linewidth]{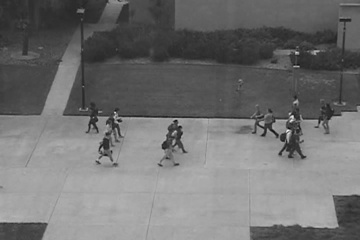}} &
\subfloat[][]{\includegraphics[width=0.48\linewidth]{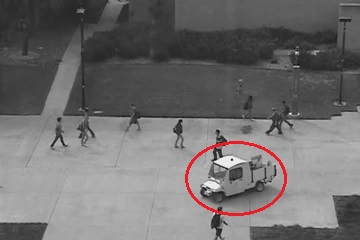}} \vspace{-0.35cm} \\
\subfloat[][]{\includegraphics[width=0.48\linewidth]{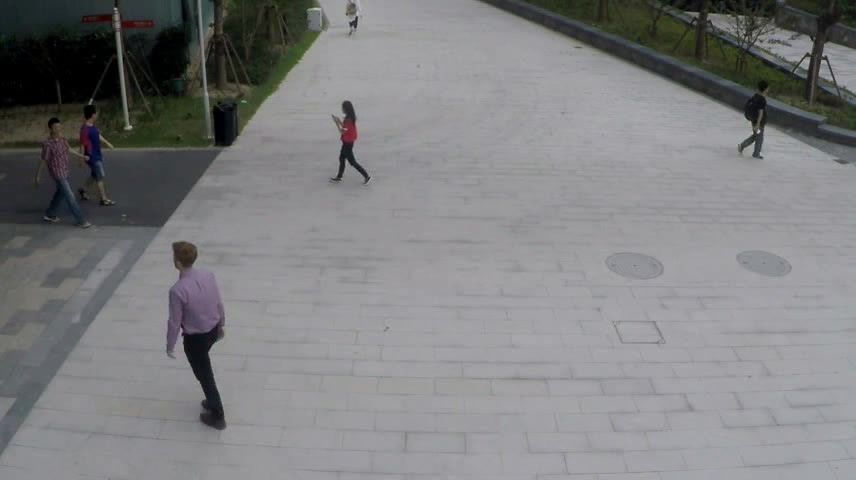}} &
\subfloat[][]{\includegraphics[width=0.48\linewidth]{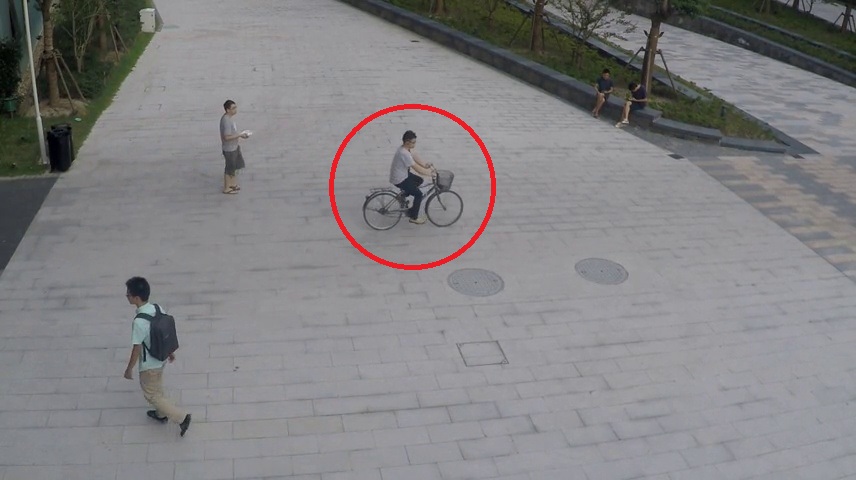}}
\end{tabular}
\caption{(a) and (c) are examples of normal scenes in UCSD Ped2 and ShanghaiTech respectively. (b) and (d) are the corresponding anomalous examples. (b) is anomalous due to the presence of a vehicle and similarly (d) is anomalous because of the appearance of a bicycle. Thus, both datasets comprise of mostly visual anomalies.}
\label{fig:ped_sh_examples}
\end{figure}

\subsection{RetroTrucks}

\input{tables/retro_examples.tex}

Anomaly detection can be very useful in moving-camera scenarios such as dashcams, bodycams, and embodied vision scenarios such as robot mounted cameras. However, to the best of our knowledge, all of the contemporary anomaly detection datasets are based on static scenes. Moreover, majority of anomalies in these datasets are caused by visual artefacts which have never been seen during training. For example, as shown in Fig. \ref{fig:ped_sh_examples}, UCSD Ped2 treats the appearance of a car as an anomaly and there is no car shown during training. This makes reconstruction-based methods a better fit as such methods struggle to reconstruct visual artefacts not seen during training. On the contrary, most real-world anomalies occur in situations where already seen objects are interacting in an abnormal manner, e.g. a car accident. This calls for reasoning beyond the visual appearance of the data. Such reasoning has been prevalent in many problems such as action recognition, video classification, etc. However, no literature exists on incorporating such reasoning in anomaly detection. We, therefore, present a new and large-scale dataset of truck dashcam videos, curated from YouTube to motivate the development of anomaly detection methods that reason beyond the visual appearance of the scenes, especially in moving-camera scenarios. Our dataset is particularly different in mainly three aspects:

\begin{itemize}
    \item All the videos are recorded from truck mounted dashcams. This introduces a new challenging camera viewpoint which has not been explored in any other dataset.
    \item All the anomalies involve the ego-vehicle. This is in contrast with existing dashcam datasets, e.g. \cite{Chan16accv}, which only show accidents between other traffic participants. 
    \item All videos are relatively long, ranging from 7 seconds to 2 minutes. 
\end{itemize}
Our dataset\footnote{RetroTrucks is available at \url{https://drive.google.com/open?id=1VxFG1jHBiep4R3i_MmvMfKWH11AEFFhu}} includes abnormal driving scenarios such as collisions, near-misses, road departures, and vehicle rollovers. Any driving scene which does not include such a scenario is included in the normal set. The driving scenes encompass a diverse set of weather and lighting conditions including rural and urban, day and night, and sunny and overcast scenes. A few examples of the driving scenes included in the dataset are illustrated in Fig. \ref{fig:retro_examples}.

We follow standard conventions from UCSD Ped2 and ShanghaiTech to organize our dataset. We train using only normal driving scenes and test on abnormal driving scenes.

\noindent \textbf{Data Collection:} We downloaded the videos from YouTube, and spliced them into normal and anomalous. We collected 474 videos, of which, 254 are normal (our training set) and 220 are anomalous. Further, 56 of the anomalous videos are annotated with temporal localization of the accidents in the videos and are used as our testing set, while the rest are provided to enable the development of weakly-supervised approaches which use both normal and anomalous videos at training. All videos have 25 FPS.

%% file: tables/retro_examples.tex
\begin{figure*}[ht!]
\centering
\setlength{\tabcolsep}{2pt}
\begin{tabular}{cc}
\subfloat[][]{\includegraphics[width=8.07cm]{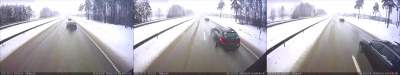}} &
\subfloat[][]{\includegraphics[width=8.2cm]{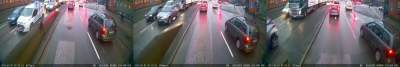}}  \vspace{-0.35cm} \\
\subfloat[][]{\includegraphics[width=8.07cm]{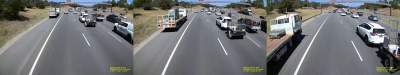}} &
\subfloat[][]{\includegraphics[width=8.2cm]{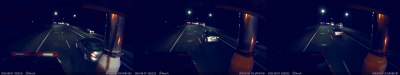}}  \vspace{-0.35cm} \\
\subfloat[][]{\includegraphics[width=8.07cm]{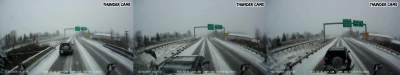}} &
\subfloat[][]{\includegraphics[width=8.2cm]{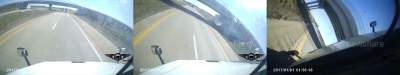}}  \vspace{-0.35cm} \\
\subfloat[][]{\includegraphics[width=8.07cm]{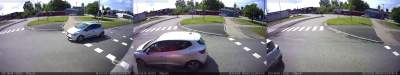}} &
\subfloat[][]{\includegraphics[width=8.2cm]{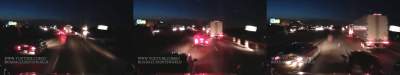}} 
\end{tabular}
\caption{Example scenes from our dataset RetroTrucks: (a)-(d) show the varying weather and lighting conditions in our dataset including snow (a), overcast (b), sunny (c) and night (d). (e)-(h) show a diverse set of anomalous examples including collision in snowy conditions (e), ego-vehicle rollover (f), near-miss (g) and collision at night time (h).}
\label{fig:retro_examples}
\end{figure*}

%% file: experiments.tex
\section{Experiments}
\label{sec:experiments}

Here, we test our one-class classification and reconstruction-based approaches on RetroTrucks. We also evaluate our object interaction modeling. These methods are also evaluated on popular anomaly detection datasets, i.e. UCSD Ped1, UCSD Ped2, and ShanghaiTech, to show a comparison between the performance of these methods on datasets with static background and on RetroTrucks. We find that although one-class classification is an interesting approach, it achieves sub-optimal performance on all the datasets we experiment with. On the other hand, reconstruction-based approaches achieve decent performance on all the datasets, including RetroTrucks. However, we will discuss some caveats of using reconstruction-based loss for real-world anomaly detection.

\subsection{One-Class Classification Approach}
\label{sec:exp_oc}

\subsubsection{Implementation Details}

We implement our one-class classification approach (Sec. \ref{sec:OC-SVDD}) using the 3D CNN architecture known as inflated 3D convnet proposed based on Resnet-50 in \cite{wang2018non}, namely \emph{I3D}. The input to I3D is of dimensions $32\times 224 \times 224 \times 3$ and it outputs a feature map of dimensions $4\times7\times7\times2048$. We add an average pooling layer followed by a linear layer to map the output to a $Z$-dimensional feature vector as discussed in Sec. \ref{sec:OC-SVDD}. We train the network using the Adam optimizer with a learning rate of $10^{-4}$ and mini-batches of size 4. The rest of the training procedure remain the same as in \cite{ruff2018deep}.

We also experiment with incorporating object interaction modeling as described in Sec. \ref{sec:gcn}. Specifically, we use a pre-trained RPN from \cite{ren2015faster}. We extract 25 object proposals for each feature frame of the bottleneck representation. Using RoIAlign followed by max pooling, we get the object features of dimensions $(4 \times 25) \times2048$. These features are then fed into a GCN which reasons for object interactions and outputs features with dimensions $(4 \times 25) \times2048$. We apply average pooling to get features with dimensions $1\times2048$. Similarly, we also apply average pooling on the output of I3D to get features with dimensions $1\times2048$. Both the above features are then concatenated to get the combined representation. We then use a linear layer to map the combined representation to a $Z$-dimensional feature vector, which is fed into the one-class SVDD objective explained in Sec. \ref{sec:OC-SVDD}. We denote this approach as \emph{I3D+GCN}. 

\subsubsection{Evaluation}

The trained network is used to map 32-frame clips to the $\mathbb{R}^{Z}$ space, where $Z = 128$. This embedding is then used to calculate the $L_2$ distance from the center $c$. We treat this distance as the anomaly score. We use the trained network in a sliding-window fashion to get anomaly scores for all 32-frame clips corresponding to each frame (16 from each side). Following \cite{luo2017revisit}, we normalize the anomaly scores for each video as below:

\vspace{-0.1cm}
\begin{equation}
\label{eq:norm}
    s_{i} = \frac{a_{i} - \min_{i}a_i}{\max_{i}a_{i} - \min_{i}a_{i}},
\end{equation}
where $a_{i}$ is the anomaly score of the $i^{th}$ frame. We then use the frame-wise AUC-ROC metric as an evaluation criterion.

\subsubsection{Results}

The results for one-class classification on video anomaly detection are given in Tab. \ref{tab:oc_results}. We note that one-class classification gives sub-optimal performance across all datasets. The I3D architecture trained with one-class classification loss achieves $0.546$, $0.543$, and $0.536$ AUC-ROC on UCSD Ped1, ShanghaiTech, and RetroTrucks respectively. This may be because the method cannot learn discriminative/representative features. Since, for one-class classification, the objective is to learn common features of variation across all videos in the training set. But, there is no constraint for what the common features could be. However, adding GCN gives a performance gain of $0.009$, $0.053$, and $0.011$ on the three datasets respectively. This shows that the object interaction modeling is useful for anomaly detection.
\input{tables/oc_results.tex}

\subsection{Reconstruction-Based Approach}
\label{sec:exp_recon}

\subsubsection{Implementation Details}

We adapt the 3D CNN in \cite{wang2018non}, i.e. \emph{I3D}, as an encoder and design a decoder with a similar architecture. The encoder takes an input clip of dimensions $32\times224\times224\times3$ and produces the bottleneck representation of dimensions $4\times7\times7\times2048$. All the experiments are performed using a mini-batch size of 4 and the Adam optimizer with a learning rate of $10^{-4}$. 

We also incorporate object interaction modeling, and find encouraging results in line with those seen for one-class classification. As done in the previous section, we use RPN and ROIAlign followed by GCN to reason for object-object relationships. The GCN outputs features with dimensions $(4\times25)\times2048$. We apply average pooling to get $1\times2048$ features. Let these features be represented by $f$. We then use two linear layers of $2048 \times 7$ neurons each to get $f^{1} = fW_{1}$ and $f^{2} = fW_{2}$, each with dimensions $1\times7\times2048$. We then use outer product of $f^{1}$ and $f^{2}$ to get $1\times7\times7\times2048$ features and then repeat $4$ times (since there are 4 feature frames in the bottleneck representation) to get $4\times7\times7\times2048$ features. The output from the outer product and the output of the encoder are then concatenated to get the combined representation, which is fed into the decoder to reconstruct the input. We denote this approach as \emph{I3D+GCN}.

\subsubsection{Evaluation}

We use the trained autoencoder to reconstruct 32-frame clips of a video in a sliding-window fashion. We use the $L_{2}$ distance between the input $x$ and the reconstruction $\hat{x}$ as the anomaly score. We then normalize the anomaly scores for each video as in Eq. \ref{eq:norm} and similarly use the frame-wise AUC-ROC as an evaluation criterion.

\subsubsection{Results}

\input{tables/main_results.tex}

\begin{figure}[ht!]
\begin{minipage}[b]{1.0\linewidth}
\vspace{-1 mm}
\centering
\resizebox{1.0\linewidth}{!}{%
\begin{tabular}{ccc}
\subfloat[][]{\includegraphics[width=0.55\linewidth, height=0.28\linewidth]{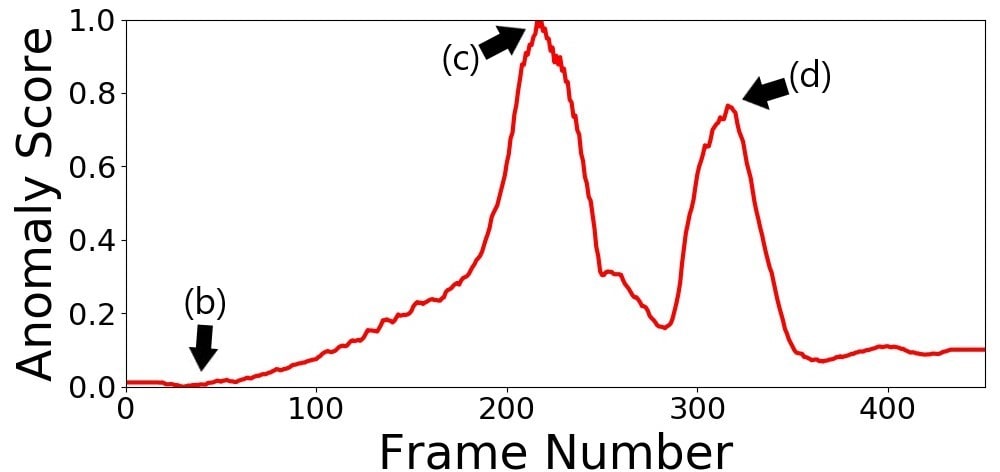}} &
\subfloat[][]{\includegraphics[width=0.5\linewidth]{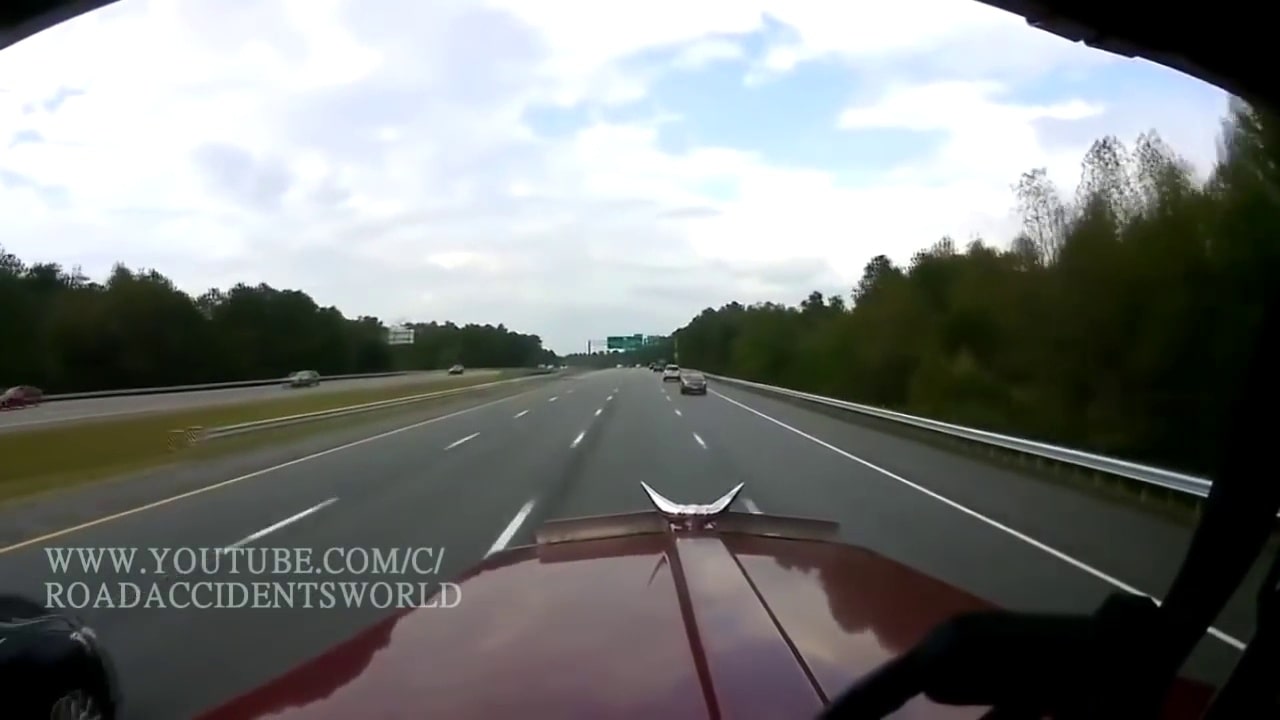}} \\
\subfloat[][]{\includegraphics[width=0.5\linewidth]{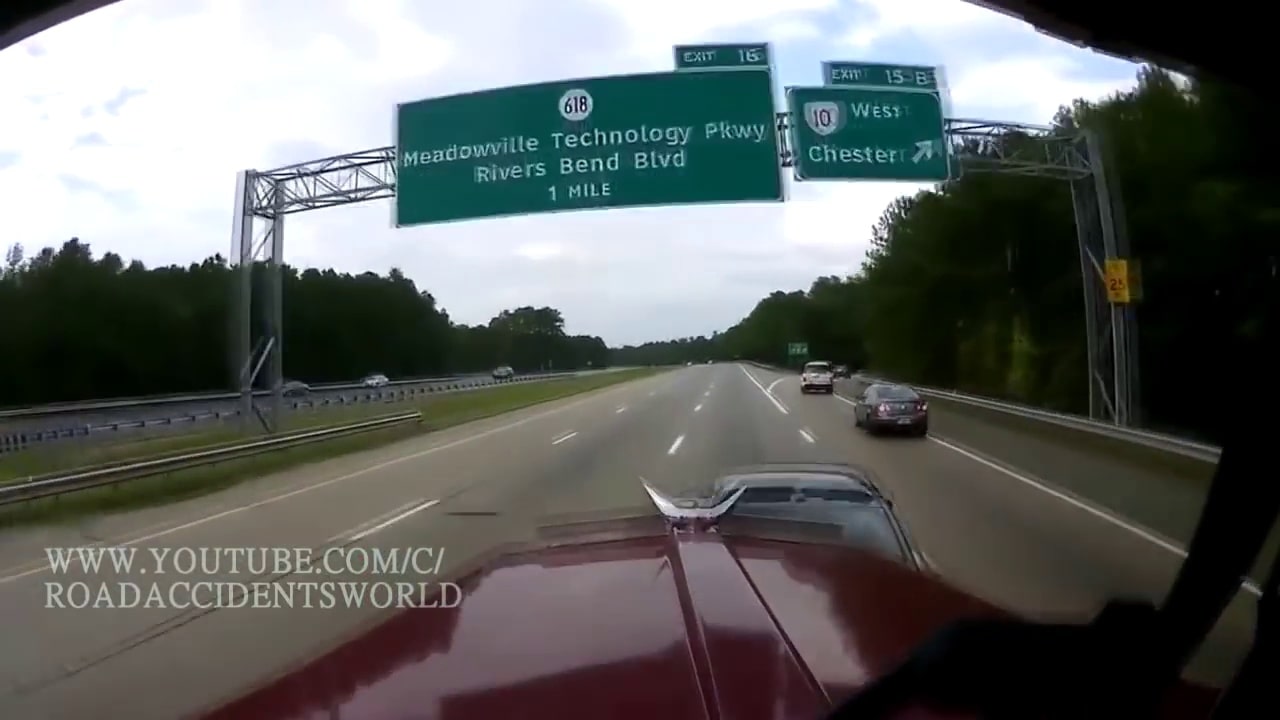}} &
\subfloat[][]{\includegraphics[width=0.5\linewidth]{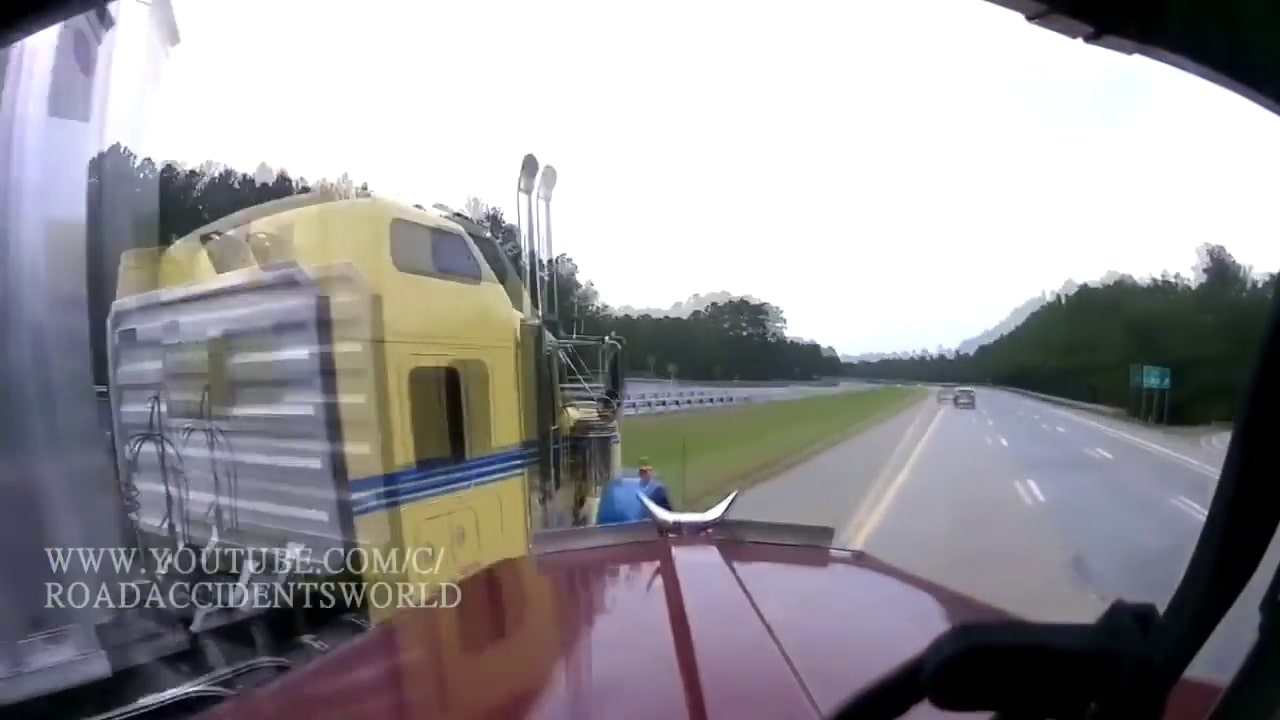}}
\end{tabular}
}
\caption{Qualitative results of I3D+GCN from Tab. \ref{tab:results}. (a) shows the predicted frame-wise anomaly scores. In (b) the ego-car is moving normally. (c) and (d) show the ego-car's collisions with another vehicle.  As is evident from (a) the model outputs high anomaly scores for the two collisions. However, it is unable to detect the skidding motion between the two collisions. We show further analysis in Sec. \ref{sec:critique}.}
\vspace{-0.3cm}
\label{fig:recon_qual_ex}
\end{minipage}
\end{figure}

The results of reconstruction-based methods are summarized in Tab. \ref{tab:results}. The best result on RetroTrucks, $0.715$ AUC-ROC, is achieved by I3D+GCN. However, adding GCN for object interaction modeling only gives an improvement of $0.003$ as opposed to an improvement of $0.011$ in Sec. \ref{sec:exp_oc}. We show an example of the qualitative performance of I3D+GCN in Fig. \ref{fig:recon_qual_ex}\footnote{Supplementary video is at: \url{https://youtu.be/AplU3JC6bjA}}. We also evaluate two state-of-the-art reconstruction-based methods on our dataset, i.e. \cite{liu2018future} and \cite{gong2019memorizing}. Both these methods perform well on UCSD Peds2, with $0.954$ and $0.941$ AUC-ROC respectively. Similarly, on ShanghaiTech both achieve $0.728$ and $0.712$ AUC-ROC respectively. However, they give worse results on our dataset, $0.606$ and $0.636$ AUC-ROC respectively. This shows that due to moving camera and contextual anomalies RetroTrucks is more challenging than the current anomaly detection datasets. 

\subsection{Ablation Study}

We also perform an ablation study to show the importance of each component of our model. Specifically, we compare the performance of the followings:
\begin{itemize}
 
  \item I3D: This is the reconstruction-based model described in Sec. \ref{sec:rec_ae}.
  \item I3D + GCN: We combine the I3D model with GCN as presented in Sec. \ref{sec:gcn}.
  
  \item I3D + Flow:  We concatenate RGB images with optical flow in the channel dimension and train a joint autoencoder which simultaneously reconstructs RGB images and optical flow. 
  \item I3D + Flow + GCN:  We add GCN on top of the I3D + Flow model described above.
\end{itemize}

The results of the ablation study are summarized in Tab. \ref{tab:ablation}. The results show that the autoencoder with I3D achieves a decent performance of $0.712$ AUC-ROC on RetroTrucks. Further, we achieve a marginal improvement of $0.003$ by adding GCN for object interaction modeling. We also experiment with optical flow, which is generally helpful for other parallel tasks such as video recognition \cite{carreira2017quo}. However, we find no performance improvement with optical flow as  I3D+Flow  achieves a lower AUC-ROC of $0.699$. 


\input{tables/ablation.tex}

\begin{figure}[ht!]
\begin{minipage}[b]{1.0\linewidth}
\centering
\resizebox{1.0\linewidth}{!}{%
\begin{tabular}{ccc}
\subfloat[][]{\includegraphics[width=0.4\linewidth]{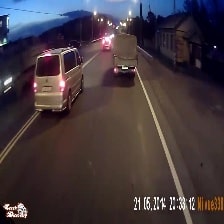}} &
\subfloat[][]{\includegraphics[width=0.4\linewidth]{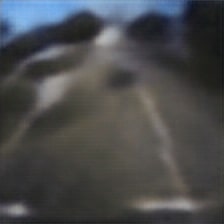}} &
\subfloat[][]{\includegraphics[width=0.4\linewidth]{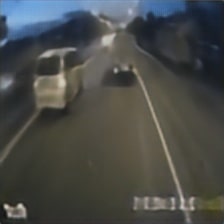}}

\end{tabular}
}
\caption{Effect of increasing bottleneck size for reconstruction-based methods. (a) is the input image. (b) is the reconstruction generated with bottleneck size of $4\times7\times7\times2048$. (c) is the reconstruction generated with bottleneck spatially doubled to $8\times14\times14\times2048$.} 
\label{fig:recon_critique}
\vspace{-0.1cm}
\end{minipage}
\end{figure}

\begin{figure}[ht!]
\vspace{-4mm}
\begin{minipage}[b]{1.0\linewidth}
\centering
\resizebox{1.0\linewidth}{!}{%
\begin{tabular}{ccc}
\multicolumn{2}{c}{\subfloat[][]{\includegraphics[width=1.0\linewidth]{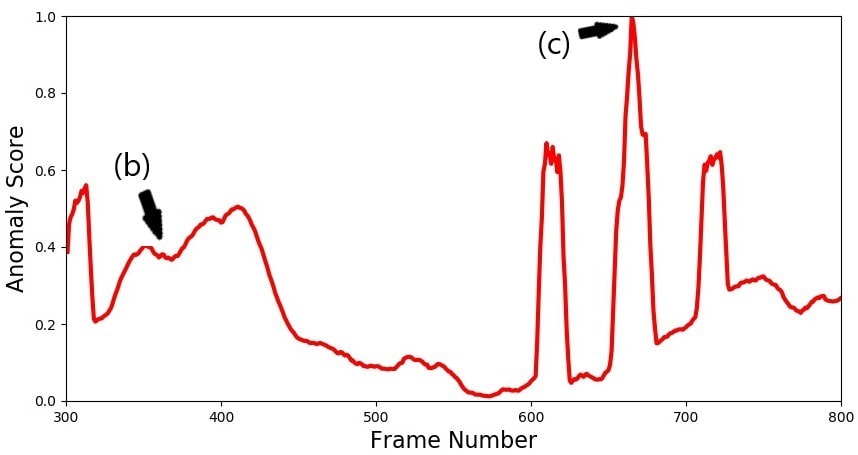}}} \\
\subfloat[][]{\includegraphics[width=0.5\linewidth, height=0.4 \linewidth]{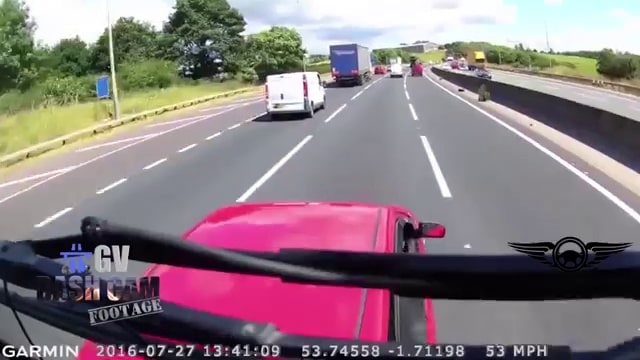}} &
\subfloat[][]{\includegraphics[width=0.5\linewidth, height=0.4 \linewidth]{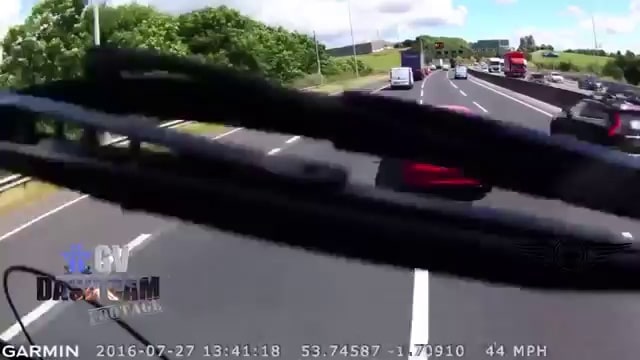}} &
\end{tabular}
}
\caption{Qualitative analysis of reconstruction-based methods. (a) is the frame-wise anomaly scores. (b) shows a car colliding with the ego-car. (c) shows a visual aberration, i.e. the camera is occluded by the front wiper of the ego car. As shown in (a), the model gives a higher anomaly score to (c) but a lower anomaly score to (b).}
\label{fig:recon_flaw}
\vspace{-0.5cm}
\end{minipage}
\end{figure}

\subsection{Discussion of Reconstruction-Based Methods} 
\label{sec:critique}

In this section, we summarize the results of reconstruction-based methods and also discuss a few failure cases. We get decent anomaly detection performance of $0.712$ AUC-ROC on RetroTrucks using an I3D-based autoencoder. However, we could not replicate the good results of reconstruction-based methods on other datasets including UCSD Ped2 and ShanghaiTech. We also add optical flow as input along with RGB frames but do not achieve any boost in performance.


However, we notice two caveats of using reconstruction-based methods for anomaly detection. Firstly, the reconstruction-based objective is not aligned with the evaluation metric for anomaly detection, i.e. better reconstruction performance does not translate to better anomaly detection performance. As we can see in Tab. \ref{tab:crit_results}, the network having bad reconstruction performance, i.e. I3D-A, has much better anomaly detection performance on RetroTrucks. An example of the reconstruction prowess of both networks can be seen in Fig. \ref{fig:recon_critique}. I3D-A has much better anomaly detection performance but mediocre reconstruction performance. On the other hand, I3D-B can reconstruct the input very well but has significantly lower anomaly detection performance. Secondly, reconstruction-based approaches can accurately detect visual anomalies, i.e. artefacts which are absent/rare in the training data, however, they fail on contextual anomalies. One example of this behavior can be seen in Fig. \ref{fig:recon_flaw}, where the model gives a higher anomaly score on the appearance of a wiper than that of the actual collision in the clip.

\begin{table}[h]
\begin{minipage}[b]{1.0\linewidth}
\centering
\small
\begin{tabular}{|l|cc|}
\hline
Network & AUC-ROC & Rec. Error \\

\hline

I3D-A (Bad Rec.) & 0.712 & 0.344 \\

I3D-B (Good Rec.)&  0.637 & 0.244 \\

\hline
\end{tabular}%
\caption{Effect of improving reconstruction on RetroTrucks. I3D-A is the same model as I3D from Tab. \ref{tab:results} whereas I3D-B has bottleneck representation spatially doubled to $8 \times 14 \times 14 \times 2048$. `Rec.' stands for reconstruction.}
\label{tab:crit_results}
\end{minipage}
\vspace{-0.7cm}
\end{table}

%% file: tables/oc_results.tex
\begin{table}[h]
\begin{minipage}[b]{1.0\linewidth}

\resizebox{0.92\textwidth}{!}{%
\begin{tabular}{|l|cc|c|}
\hline
Method & UCSD Ped1 & ShanghaiTech & RetroTrucks  \\

\hline

I3D & 0.546 & 0.543 & 0.536 \\

I3D + GCN & 0.555 &  0.596 &  0.547 \\

\hline
\end{tabular}%
}
\caption{AUC-ROC of one-class classification methods on UCSD Ped1, ShanghaiTech, and RetroTrucks.}
\label{tab:oc_results}
\end{minipage}

\end{table}

%% file: tables/main_results.tex

\begin{table*}
\vspace{0.2cm}
\begin{minipage}[b]{1.0\linewidth}
\centering

\resizebox{0.82\textwidth}{!}{%
\begin{tiny}
\begin{tabular}{|l|l|ccc|c|}

\hline

&Method & UCSD Ped1 & UCSD Ped2 & ShanghaiTech & RetroTrucks  \\
\hline

\parbox[t]{2mm}{\multirow{3}{*}{\rotatebox[origin=c]{90}{\tiny{Non-Rec.}}}}

& Unmasking
\cite{tudor2017unmasking} 
& 0.684 & 0.822 & - & - \\

& AMDN
\cite{xu2015AMDN}
& 0.921 & 0.908 & - & - \\

& FutureFrame 
~\cite{liu2018future} 
& 0.831 & 0.954 & 0.728 & 0.606 \\


\hline
\parbox[t]{2mm}{\multirow{3}{*}{\rotatebox[origin=c]{90}{\tiny{Rec.}}}}
& MemAE
~\cite{gong2019memorizing} 
&  - & 0.941 & 0.712 & 0.636 \\


& AbnormalGAN
~\cite{ravanbakhsh2017abnormal} 
& 0.974 & 0.935 & -  & - \\

& AE-3D (MemAE)
~\cite{gong2019memorizing} 
& - & 0.921 & 0.701 & 0.640 \\

\hline

& I3D
& 0.732 & 0.700   & 0.689 &  0.712 \\
& I3D + GCN
&  0.740   & 0.693   & 0.694  &  0.715  \\

\hline

\end{tabular}%
\end{tiny}
}
\caption{AUC-ROC of different methods on UCSD Ped1, UCSD Ped2, ShanghaiTech, and RetroTrucks. `Rec.' denotes reconstruction-based methods, while `Non-Rec.' represents all other methods which do not have a reconstruction-based loss.}
\label{tab:results}
\vspace{-0.15cm}
\end{minipage}

\end{table*}

%% file: tables/ablation.tex

\begin{table}
\hspace{2mm}
\begin{minipage}[b]{1.0\linewidth}

\centering
\resizebox{0.58\textwidth}{!}{%
\begin{tabular}{|l|c|}
\hline
Method & AUC-ROC  \\

\hline
I3D & 0.712 \\

I3D + GCN & 0.715 \\



I3D + Flow & 0.699 \\

I3D + Flow + GCN & 0.696 \\

\hline
\end{tabular}
}
\caption{Ablation of our methods on RetroTrucks.}
\label{tab:ablation}
\vspace{-0.1cm}
\end{minipage}
\end{table}

%% file: conclusion.tex
\section{Conclusion}
\label{sec:conclusion}

We present a new challenging dataset of anomaly detection in dashcam videos with a diverse set of accidents and road scenes. We evaluate our new dataset with two data-driven approaches, i.e. one-class classification and reconstruction-based. The experimental results show that although reconstruction-based methods work well for datasets in which anomalous examples are characterized by visual artefacts, they do not perform as well on our dataset since the anomalies are characterized by contextual information rather than visual aberrations. Moreover, we also experiment with feature representations for modelling object interactions and observe improvement in the performance. Our future work will explore more effective techniques for RGB images and optical flow fusion, e.g. cross-channel fusion~\cite{ravanbakhsh2019training}, or deep supervision, e.g. object detection as an intermediate task~\cite{li2018deep,fathy2018hierarchical}. Another direction for future work is to remove visual artefacts, e.g. radial distortion~\cite{zhuang2019degeneracy}, rolling shutter effect~\cite{zhuang2019learning}, and text~\cite{hertz2019blind}, from the input video before processing.


%% file: root.bbl
\begin{thebibliography}{10}
\providecommand{\url}[1]{#1}
\csname url@rmstyle\endcsname
\providecommand{\newblock}{\relax}
\providecommand{\bibinfo}[2]{#2}
\providecommand\BIBentrySTDinterwordspacing{\spaceskip=0pt\relax}
\providecommand\BIBentryALTinterwordstretchfactor{4}
\providecommand\BIBentryALTinterwordspacing{\spaceskip=\fontdimen2\font plus
\BIBentryALTinterwordstretchfactor\fontdimen3\font minus
  \fontdimen4\font\relax}
\providecommand\BIBforeignlanguage[2]{{%
\expandafter\ifx\csname l@#1\endcsname\relax
\typeout{** WARNING: IEEEtran.bst: No hyphenation pattern has been}%
\typeout{** loaded for the language `#1'. Using the pattern for}%
\typeout{** the default language instead.}%
\else
\language=\csname l@#1\endcsname
\fi
#2}}

\bibitem{mukhtar2015vehicle}
A.~Mukhtar, L.~Xia, and T.~B. Tang, ``Vehicle detection techniques for
  collision avoidance systems: A review,'' \emph{T-ITS}, 2015.

\bibitem{narote2018review}
S.~P. Narote, P.~N. Bhujbal, A.~S. Narote, and D.~M. Dhane, ``A review of
  recent advances in lane detection and departure warning system,'' \emph{PR},
  2018.

\bibitem{liu2017radar}
G.~Liu, L.~Wang, and S.~Zou, ``A radar-based blind spot detection and warning
  system for driver assistance,'' in \emph{IAEAC}, 2017.

\bibitem{schneider2019lidar}
K.~Schneider, R.~Lugner, and T.~Brandmeier, ``Lidar-based contour estimation of
  oncoming vehicles in pre-crash scenarios,'' in \emph{IV}, 2019.

\bibitem{karpathy2014large}
A.~Karpathy, G.~Toderici, S.~Shetty, T.~Leung, R.~Sukthankar, and L.~Fei-Fei,
  ``Large-scale video classification with convolutional neural networks,'' in
  \emph{CVPR}, 2014.

\bibitem{carreira2017quo}
J.~Carreira and A.~Zisserman, ``Quo vadis, action recognition? a new model and
  the kinetics dataset,'' in \emph{CVPR}, 2017.

\bibitem{idrees2017thumos}
H.~Idrees, A.~R. Zamir, Y.-G. Jiang, A.~Gorban, I.~Laptev, R.~Sukthankar, and
  M.~Shah, ``The thumos challenge on action recognition for videos “in the
  wild”,'' \emph{CVIU}, 2017.

\bibitem{li2013anomaly}
W.~Li, V.~Mahadevan, and N.~Vasconcelos, ``Anomaly detection and localization
  in crowded scenes,'' \emph{T-PAMI}, 2013.

\bibitem{lu2013abnormal}
C.~Lu, J.~Shi, and J.~Jia, ``Abnormal event detection at 150 fps in matlab,''
  in \emph{ICCV}, 2013.

\bibitem{luo2017revisit}
W.~Luo, W.~Liu, and S.~Gao, ``A revisit of sparse coding based anomaly
  detection in stacked rnn framework,'' in \emph{ICCV}, 2017.

\bibitem{Sultani2018RealWorldAD}
W.~Sultani, C.~Chen, and M.~Shah, ``Real-world anomaly detection in
  surveillance videos,'' in \emph{CVPR}, 2018.

\bibitem{khan2014one}
S.~S. Khan and M.~G. Madden, ``One-class classification: taxonomy of study and
  review of techniques,'' \emph{The Knowledge Engineering Review}, 2014.

\bibitem{ruff2018deep}
L.~Ruff, R.~Vandermeulen, N.~Goernitz, L.~Deecke, S.~A. Siddiqui, A.~Binder,
  E.~M{\"u}ller, and M.~Kloft, ``Deep one-class classification,'' in
  \emph{ICML}, 2018.

\bibitem{perera2019learning}
P.~Perera and V.~M. Patel, ``Learning deep features for one-class
  classification,'' \emph{T-IP}, 2019.

\bibitem{an2015variational}
J.~An and S.~Cho, ``Variational autoencoder based anomaly detection using
  reconstruction probability,'' \emph{Special Lecture on IE}, 2015.

\bibitem{zhao2017spatio}
Y.~Zhao, B.~Deng, C.~Shen, Y.~Liu, H.~Lu, and X.-S. Hua, ``Spatio-temporal
  autoencoder for video anomaly detection,'' in \emph{MM}, 2017.

\bibitem{liu2018future}
W.~Liu, W.~Luo, D.~Lian, and S.~Gao, ``Future frame prediction for anomaly
  detection--a new baseline,'' in \emph{CVPR}, 2018.

\bibitem{gong2019memorizing}
D.~Gong, L.~Liu, V.~Le, B.~Saha, M.~R. Mansour, S.~Venkatesh, and A.~v.~d.
  Hengel, ``Memorizing normality to detect anomaly: Memory-augmented deep
  autoencoder for unsupervised anomaly detection,'' in \emph{ICCV}, 2019.

\bibitem{dhiman2016continuous}
V.~Dhiman, Q.-H. Tran, J.~J. Corso, and M.~Chandraker, ``A continuous occlusion
  model for road scene understanding,'' in \emph{CVPR}, 2016.

\bibitem{li2017deep}
C.~Li, M.~Zeeshan~Zia, Q.-H. Tran, X.~Yu, G.~D. Hager, and M.~Chandraker,
  ``Deep supervision with shape concepts for occlusion-aware 3d object
  parsing,'' in \emph{CVPR}, 2017.

\bibitem{song2017fcwstereo}
W.~Song, M.~Fu, Y.~Yang, M.~Wang, X.~Wang, and A.~Kornhauser, ``Real-time lane
  detection and forward collision warning system based on stereo vision,'' in
  \emph{IV}, 2017.

\bibitem{matousek2019detecting}
M.~Matousek, E.-Z. Mohamed, F.~Kargl, C.~B{\"o}sch, \emph{et~al.}, ``Detecting
  anomalous driving behavior using neural networks,'' in \emph{IV}, 2019.

\bibitem{fang2019dada}
J.~Fang, D.~Yan, J.~Qiao, J.~Xue, H.~Wang, and S.~Li, ``Dada-2000: Can driving
  accident be predicted by driver attention analyzed by a benchmark,'' in
  \emph{ITSC}, 2019.

\bibitem{wang2018videos}
X.~Wang and A.~Gupta, ``Videos as space-time region graphs,'' in \emph{ECCV},
  2018.

\bibitem{ren2015faster}
S.~Ren, K.~He, R.~Girshick, and J.~Sun, ``Faster r-cnn: Towards real-time
  object detection with region proposal networks,'' in \emph{NeurIPS}, 2015.

\bibitem{he2017mask}
K.~He, G.~Gkioxari, P.~Doll{\'a}r, and R.~Girshick, ``Mask r-cnn,'' in
  \emph{ICCV}, 2017.

\bibitem{kipf2016semi}
T.~N. Kipf and M.~Welling, ``Semi-supervised classification with graph
  convolutional networks,'' in \emph{ICLR}, 2017.

\bibitem{herzig2019spatio}
R.~Herzig, E.~Levi, H.~Xu, H.~Gao, E.~Brosh, X.~Wang, A.~Globerson, and
  T.~Darrell, ``Spatio-temporal action graph networks,'' in \emph{ICCVW}, 2019.

\bibitem{Chan16accv}
F.-H. Chan, Y.-T. Chen, Y.~Xiang, and M.~Sun, ``Anticipating accidents in
  dashcam videos,'' in \emph{ACCV}, 2016.

\bibitem{che2019d}
Z.~Che, G.~Li, T.~Li, B.~Jiang, X.~Shi, X.~Zhang, Y.~Lu, G.~Wu, Y.~Liu, and
  J.~Ye, ``D$^2$-city: A large-scale dashcam video dataset of diverse traffic
  scenarios,'' \emph{arXiv}, 2019.

\bibitem{wang2018non}
X.~Wang, R.~Girshick, A.~Gupta, and K.~He, ``Non-local neural networks,'' in
  \emph{CVPR}, 2018.

\bibitem{tudor2017unmasking}
R.~Tudor~Ionescu, S.~Smeureanu, B.~Alexe, and M.~Popescu, ``Unmasking the
  abnormal events in video,'' in \emph{ICCV}, 2017.

\bibitem{xu2015AMDN}
D.~Xu, E.~Ricci, Y.~Yan, J.~Song, and N.~Sebe, ``Learning deep representations
  of appearance and motion for anomalous event detection,'' in \emph{BMVC},
  2015.

\bibitem{ravanbakhsh2017abnormal}
M.~Ravanbakhsh, M.~Nabi, E.~Sangineto, L.~Marcenaro, C.~Regazzoni, and N.~Sebe,
  ``Abnormal event detection in videos using generative adversarial nets,'' in
  \emph{ICIP}, 2017.

\bibitem{ravanbakhsh2019training}
M.~Ravanbakhsh, E.~Sangineto, M.~Nabi, and N.~Sebe, ``Training adversarial
  discriminators for cross-channel abnormal event detection in crowds,'' in
  \emph{WACV}, 2019.

\bibitem{li2018deep}
C.~Li, M.~Z. Zia, Q.-H. Tran, X.~Yu, G.~D. Hager, and M.~Chandraker, ``Deep
  supervision with intermediate concepts,'' \emph{T-PAMI}, 2018.

\bibitem{fathy2018hierarchical}
M.~E. Fathy, Q.-H. Tran, M.~Zeeshan~Zia, P.~Vernaza, and M.~Chandraker,
  ``Hierarchical metric learning and matching for 2d and 3d geometric
  correspondences,'' in \emph{ECCV}, 2018.

\bibitem{zhuang2019degeneracy}
B.~Zhuang, Q.-H. Tran, G.~H. Lee, L.~F. Cheong, and M.~Chandraker, ``Degeneracy
  in self-calibration revisited and a deep learning solution for uncalibrated
  slam,'' in \emph{IROS}, 2019.

\bibitem{zhuang2019learning}
B.~Zhuang, Q.-H. Tran, P.~Ji, L.-F. Cheong, and M.~Chandraker, ``Learning
  structure-and-motion-aware rolling shutter correction,'' in \emph{CVPR},
  2019.

\bibitem{hertz2019blind}
A.~Hertz, S.~Fogel, R.~Hanocka, R.~Giryes, and D.~Cohen-Or, ``Blind visual
  motif removal from a single image,'' in \emph{CVPR}, 2019.

\end{thebibliography}
